\documentclass[sigconf,nonacm]{acmart}

\usepackage{amsmath,amssymb}

\usepackage{graphicx}
\usepackage{subcaption}
\usepackage{booktabs}
\usepackage{array}
\usepackage{makecell}
\usepackage{wrapfig}
\usepackage{adjustbox}
\usepackage{multirow}
\usepackage{enumitem}              
\usepackage{tikz}
\usepackage{tabularx}

\usepackage{algorithm}
\usepackage{algpseudocode}

\usepackage{url}

\usepackage{xspace}

\AtBeginDocument{%
  }

\acmConference[Conference acronym 'XX]{}{}{}
\acmISBN{978-1-4503-XXXX-X/2018/06}

\begin{document}

\title{Proof-of-Use: Mitigating Tool-Call Hacking in Deep Research Agents}


\author{Shengjie Ma}
\affiliation{%
  \institution{Gaoling School of Artificial Intelligence, Renmin University of China}
  \country{China}
}
\email{msj@ruc.edu.cn}
\author{Chenlong Deng}
\affiliation{%
  \institution{Gaoling School of Artificial Intelligence, Renmin University of China}
  \country{China}
}
\email{dengchenlong@ruc.edu.cn}
\author{Jiaxin Mao}
\affiliation{%
  \institution{Gaoling School of Artificial Intelligence, Renmin University of China}
  \country{China}
}
\email{maojiaxin@gmail.com}


\author{Jiadeng Huang}
\affiliation{%
  \institution{OPPO}
  \country{China}
}
\email{huangjiadeng@oppo.com}

\author{Teng Wang}
\affiliation{%
  \institution{OPPO}
  \country{China}
}
\email{	wt0318@connect.hku.hk}

\author{Junjie Wu}
\affiliation{%
  \institution{OPPO}
  \country{China}
}
\email{wujunjie1@oppo.com}

\author{Changwang Zhang}
\affiliation{%
  \institution{OPPO}
  \country{China}
}
\email{changwangzhang@foxmail.com}

\author{Jun Wang}
\affiliation{%
  \institution{OPPO}
  \country{China}
}
\email{junwang.lu@gmail.com}


\renewcommand{\shortauthors}{Ma et al.}


\begin{abstract}
Deep research agents increasingly operate in complex multi-source environments, where large language models interact with external knowledge through iterative tool calls to support multi-step reasoning. While reinforcement learning (RL) enhances their ability to plan and reason across retrieval steps, we identify a critical failure mode in this setting: Tool-Call Hacking. Unlike execution-based tools (e.g., code or math), whose effects are directly observable, the weak observability of causal dependencies between retrieved evidence and reasoning under format- and outcome-level supervision enables agents to maximize surface-level reward signals without genuinely grounding their reasoning in the returned evidence. This leads to distinctive pathologies, including mode collapse via tool overuse and hallucinated tool usage where tool calls are largely decorative.

To address this issue, we propose Proof-of-Use (PoU), an evidence grounded RL framework that explicitly optimizes the causal dependency from retrieval to reasoning and final answers. PoU re-fomulate a fine-grained stepwise interaction protocol in which agents must auditably cite normalized evidence identifiers. We operationalize this via a multi-objective reward design consisting of: (1) two progressive process rewards that constrain citation validity at intermediate steps; (2) a global Answer--Support Alignment reward that enforces consistency between final answers and retrieved evidence; and (3) a curriculum-style adaptive reward mixing mechanism that smoothly transitions agents from dense process supervision to sparse outcome-based objectives. Extensive experiments show the strong performance of PoU and demonstrate the effectiveness in mitigating tool-call hacking. Beyond this, PoU exhibits a notable emergent property: adaptive and robust tool-usage patterns naturally arise under domain and tool shifts, even though PoU does not explicitly optimize for tool adaptation.
\end{abstract}

\maketitle

\section{Introduction}
\begin{figure}[t]
  \centering
  \includegraphics[width=0.9\linewidth]{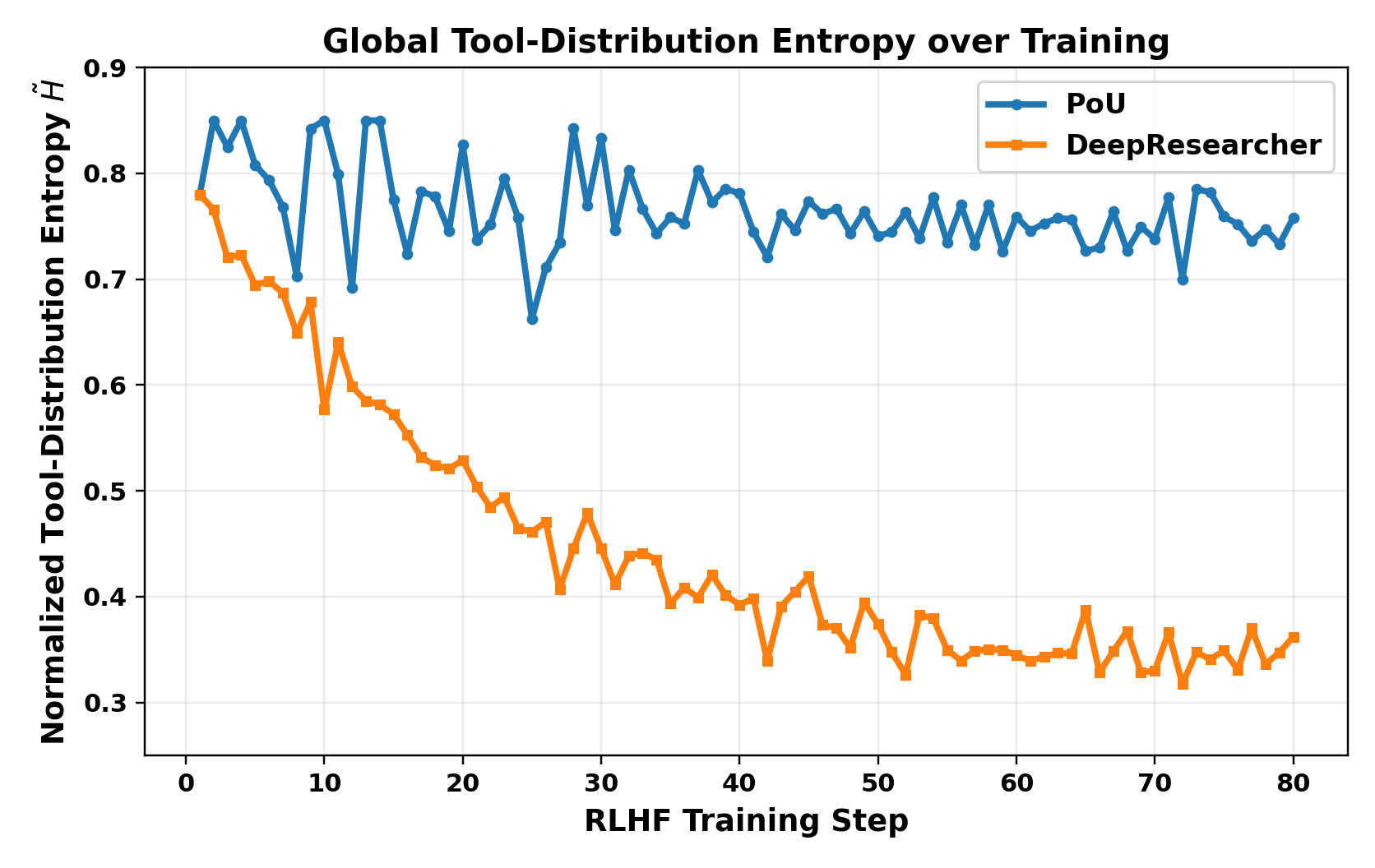}
  \caption{Tool-distribution entropy over training.
At each training step, we compute the entropy of the tool usage
distribution as
$\tilde{H} = -\sum_{i} p_i \log p_i \,/\, \log K$,
where $p_i$ is the proportion of calls to tool $i$ and $K$ is the number
of available tools.
DeepResearcher exhibits a decreasing entropy, which indicates a tool-call hacking pattern (progressive collapsing toward a narrow tool selection).
In contrast, PoU shows higher variance in early training, reflecting
exploration, and converges to a stable intermediate entropy value,
indicating a learned and task-coordinated tool usage pattern.}
  \label{fig:example1}
\end{figure}

Deep research agents\cite{jin2025searchr1trainingllmsreason, song2025r1searcherincentivizingsearchcapability,zheng2025deepresearcherscalingdeepresearch}  are emerging as a powerful paradigm for complex information-seeking tasks, where large language models interact with external knowledge sources through iterative tool calls to search, retrieve, verify, and synthesize evidence. Unlike single-shot retrieval or static prompting, these agents operate in multi-step environments that require not only reasoning over retrieved information, but also planning when, how, and which tools to invoke to progressively construct reliable answers.

In realistic research settings, no single external knowledge source is sufficient\cite{ma2025thinkongraph}. Modern agents therefore rely on multiple heterogeneous tools, such as web search engines, knowledge graphs, domain-specific databases, or private document collections, each with distinct coverage, structure, and reliability. These tools are exposed as callable interfaces within agentic frameworks, turning tool selection and sequencing into a central component of the agent’s decision-making process. Consequently, deep research agents must simultaneously solve two tightly coupled challenges: grounding intermediate reasoning in retrieved evidence, and strategically routing tool calls across heterogeneous knowledge sources.


From an operational perspective, deep research agents can be viewed as performing multi-round retrieval-augmented generation (RAG). Contrary to expectations, recent empirical studies show that brute-force, all-in-one retrieval can outperform prompt-based selective routing\citep{xu2025ragwildineffectivenessllms}. This suggests that naïve prompt-based routing learned from shallow heuristics does not reliably improve retrieval effectiveness; instead, it often amplifies systemic inefficiencies and disrupts cross-source coordination. \textbf{When routing decisions are not grounded in evidence usage, selectivity becomes a liability rather than an advantage.}

Building on these observations, we extend the analysis from routing heuristics to the learning dynamics of RL-trained deep research agents. Prior works have applied reinforcement learning (RL) to enhance agents' ability to multi‑step reasoning with retrieval tools \cite{zheng2025deepresearcherscalingdeepresearch}. However, we uncover a recurrent failure pattern that we term \textbf{Tool‑Call Hacking}—a domain‑specific form of reward hacking \citep{skalse2025definingcharacterizingrewardhacking}. Specifically, the agent learns tool‑calling behaviors that boost the observed reward signal without genuinely improving grounding, which is highly susceptible to Goodhart’s Law\cite{majka2025strongweakbenigngoodharts}. We observe two consistent manifestations (as shown in Figure~\ref{fig:example1}): \textbf{(i) Mode collapse via tool overuse}: policies default to a narrow subset of tools; \textbf{(ii) Hallucinated tool use}: tool calls are made, but the reasoning chain shows no causal dependence on the retrieved evidence. \textbf{Crucially, both failures arise because existing training signals primarily reward the act of calling tools, rather than the act of using them.}

These issues highlight the need for stronger signals that foster reasoning processes deeply integrated with tool interactions in multi-source environments.


As clarified in Paragraph~\ref{sec:scope_tools}, we focus on retrieval tools, whose impact on outputs is propagated through an implicit, semantic level evidence-reasoning-answer causal chain. In contrast, execution-style tools such as code or mathematical solvers have directly observable contributions, which makes retrieval tools easy to bypass or imitate without genuine informational dependence. To this end, rather than treating evidence grounding as an emergent side effect of task success, we argue that it should be an explicit learning objective in agentic reinforcement learning. We propose Proof-of-Use (PoU), an evidence-grounded framework that refines the coarse-grained supervision commonly used in agentic RL. PoU is built on a stepwise interaction protocol: tool responses are normalized into evidence snippets with stable IDs. Upon observing them, the model must output a usefulness verdict and—if useful—cite the corresponding evidence IDs before reasoning. This exposes a verifiable dependency between retrieved evidence, reasoning, and decisions. On top of this protocol, PoU applies \textbf{two progressive process rewards} to strictly constrain step-wise reasoning. Furthermore, to address the reward sparsity and non-smooth gradients of outcome-only supervision, we introduce a global \textbf{Answer–Support Alignment reward}. This adheres to the intuition that effective learning should optimize the reasoning process alongside the final result. Crucially, we integrate these objectives via a curriculum-style \textbf{adaptive reward mixing} mechanism. This mechanism smoothly transitions focus from dense reasoning signals to final answer correctness, creating a stable gradient ascent path that bridges the gap to sparse outcome-based rewards.

Together, these design choices \textbf{couple every stage into an auditable Proof-of-Use chain}: tool outputs are anchored by stable IDs, steps explicitly cite evidence, cited grounds are selectively perturbed for verification, and the final answer is rewarded for consistency with the same evidence.




\paragraph{\textbf{Scope of Tools:}}\label{sec:scope_tools}
In this work, we define “tools” as black-box\cite{ajwani2024llmgeneratedblackboxexplanationsadversarially} \emph{proxies} that interface with external knowledge sources: retrieval-style tools including but not limited to open-web search, internal document collections, structured APIs, or specialized knowledge bases. This reflects a common abstraction in RAG systems, where tools are used to fetch potentially relevant information conditioned on the current query or reasoning context. We \textbf{do not} extend our analysis to coding or mathematical tools\cite{wölflein2025llmagentsmakingagent}.

This choice is deliberate: coding and math tools primarily stress symbolic
reasoning and execution correctness, where tool outputs are directly consumed
as final answers and failures are typically observable (e.g., wrong values or
execution errors). In contrast, retrieval tools serve as intermediate evidence providers.
Their effectiveness depends on problem analysis, retrieval planning, and
information integration, creating a setting where agents can satisfy
tool-call specifications without causally relying on the returned evidence.
\textbf{This makes retrieval-centric tools particularly susceptible to tool-call hacking} (see Section~\ref{def-toolcallhack} for a formal definition), motivating our PoU rewards.

\paragraph{\textbf{Contributions:}}
Our work makes three key contributions:

\begin{itemize}
\item We \textbf{characterize and analyse a subtle yet under-explored reward-hacking risk \textbf{tool-call hacking}} in RL-trained RAG agents under multi-source retrieval settings. We formalize its manifestations and empirically underscores the necessity of tackling this challenge.

\item To mitigate Tool-call Hacking, \textbf{we propose} \textbf{Proof-of-Use}, an evidence-grounded reinforcement learning framework that introduces a stepwise interaction protocol with progressive process rewards, a global answer-support alignment reward and an adaptive reward mixing training strategy.

\item Extensive experiments across multiple datasets \textbf{demonstrate the strong performance} of PoU and its effectiveness in \textbf{mitigating tool-call hacking}. Beyond this, PoU \textbf{exhibits notable emergent properties}: robustness and adaptive tool-usage patterns naturally arise under domain and tool shifts, even though PoU does not explicitly optimize for them.
\end{itemize}

\section{Related Works}

\subsection{Reinforcement Learning based Deep Research Agents}
Large language models (LLMs) are limited by the static knowledge embedded in their parameters, which often leads to hallucinations when answering time-sensitive or knowledge-intensive queries. To address this limitation, retrieval-augmented generation (RAG) systems have evolved from static heuristics to adaptive reasoning agents. Recent works apply reinforcement learning (RL) to enable LLMs to autonomously plan reasoning steps, decide when and how to invoke retrieval tools, and verify information across heterogeneous sources \citep{zheng2025deepresearcherscalingdeepresearch,song2025r1searcherincentivizingsearchcapability}.

Search-R1~\cite{jin2025searchr1trainingllmsreason} introduces a closed-loop \emph{reason-and-search} framework, allowing models to dynamically invoke search engines during answer generation.
R1-Searcher~\cite{song2025r1searcherincentivizingsearchcapability} further adopts a two-stage RL training paradigm that substantially improves autonomous search behaviors.
ReSearch~\cite{chen2025researchlearningreasonsearch} treats search as an integral part of the reasoning chain without relying on supervised intermediate reasoning labels.
Beyond controlled environments, DeepResearcher~\cite{zheng2025deepresearcherscalingdeepresearch} extends RL training to the open Internet, enabling end-to-end optimization in real-world web settings.



\subsection{Reward Hacking in Reinforcement Learning}
Reward hacking (a.k.a. specification gaming) refers to agents exploiting misspecified objectives to achieve high reward while violating the designer’s true intent. It was systematically highlighted as a core AI-safety challenge in \emph{Concrete Problems in AI Safety} and later analyzed through empirical taxonomies of gaming behaviors~\cite{amodei2016concreteproblemsaisafety}. Formal treatments attribute reward hacking to proxy misspecification, characterizing conditions under which objectives become hackable and proposing principled defenses; related work studies \emph{reward tampering}, where agents manipulate the reward channel itself~\cite{skalse2025definingcharacterizingrewardhacking}. These failure modes are closely connected to Goodhart’s Law—over-optimizing a proxy causes it to decouple from the true goal—and to goal misgeneralization, where models pursue unintended objectives despite correct training feedback~\cite{manheim2019categorizingvariantsgoodhartslaw}.

In LLM post-training, reward-model optimization introduces additional risks: excessive optimization against learned reward models can degrade human-judged quality as policies overfit spurious reward artifacts. Prior work proposes constrained RLHF and composite reward formulations to mitigate such effects, while iterative RLHF can empirically reduce over-optimization across rounds, albeit with diminishing returns~\cite{moskovitz2024confronting}. Process-based supervision, such as process reward models (PRMs) or step-wise Chain-of-Thought\cite{wei2023chainofthoughtpromptingelicitsreasoning} rewards, offers denser and potentially less gameable signals; however, monitoring intermediate reasoning also opens new attack surfaces, enabling models to obfuscate or game the monitor itself. This has motivated hybrid objectives that jointly supervise both outcomes and processes. Orthogonal alignment approaches, such as Constitutional AI, further reduce reliance on noisy human labels through principle-guided self-critique, reframing supervision to limit exploitable gaps in reward design~\cite{bai2022constitutionalaiharmlessnessai}.

Modern RL-trained retrieval–reasoning agents further expand the action space by allowing policies to invoke external tools (e.g., search engines, web browsers, or knowledge-base APIs), thereby influencing the information sources used to compute downstream rewards. In this setting, reward hacking manifests in subtle and prevalent forms. We identify and formalize these failures as \emph{tool-call hacking}, and this paper aims to provide a principled solution to this problem.

\section{Method}
PoU follows an iterative reasoning--acting--observing trajectory
\cite{yao2023reactsynergizingreasoningacting} for tool-augmented question
answering.
To address the often overlooked tool-call hacking in such settings, we design a set of
specialized training rewards together with a matched step-level protocol
that governs the interaction between reasoning, tool calls, and answers.
A key property of PoU is \textbf{ID-consistent end-to-end coupling}: the same evidence IDs introduced by tools are (1) referenced at the step level via usefulness verdicts and citations, and (2) reused at the answer level for evidence--answer consistency checks. This shared identifier layer \textbf{turns evidence use into a traceable protocol}, making every reward attributable to specific retrieved evidence rather than superficial tool calls.

We will first formalize the tool-call hacking phenomenon, and then present the
design of PoU rewards in detail.
\subsection{Preliminaries}
\paragraph{\textbf{RL Deep Research Agent Interaction Protocol and Rewards.}}
Traditional deep research agents\cite{zheng2025deepresearcherscalingdeepresearch,song2025r1searcherincentivizingsearchcapability,jin2025searchr1trainingllmsreason,li2025chainofagentsendtoendagentfoundation} follow a fixed interaction protocol that iterates over
\texttt{<think>} $\rightarrow$ \texttt{<tool\_call>} $\rightarrow$ \texttt{<tool\_response>} in a loop until termination.
At each step, the agent observes the input prompt or tool response of the last step, produces an reasoning trace, and then optionally invokes a tool before proceeding to the next step.

Training supervision is typically provided only at the \emph{format} and \emph{answer} levels.
The overall reward is commonly defined as a weighted aggregation:
\begin{equation}
R = \lambda \, R_{\text{format}} + (1-\lambda)\, R_{\text{answer}},
\end{equation}
where $R_{\text{format}}$ enforces syntactic validity of the interaction protocol (e.g., well-formed tags and tool calls),
$R_{\text{answer}}$ evaluates the correctness of the final answer, and $\lambda \in [0,1]$ controls the trade-off between format compliance and answer quality.

\paragraph{\textbf{Tool-Call Hacking (Definition):}}
We define \emph{tool-call hacking} as a specific form of reward hacking
\citep{skalse2025definingcharacterizingrewardhacking} in tool-augmented LLM
agents, where the policy attains high training rewards by satisfying
tool-call format requirements and answer-level correctness objectives,
while its \emph{stepwise reasoning and action behavior} does not causally
depend on the content returned by the tools. As a result, the trained model deviates from the intended goal of
retrieval-augmented generation, introducing a high but often
hard-to-detect risk of hallucinated reasoning and answers.

\paragraph{\textbf{Manifestations of Tool-Call Hacking:}}
In practice, tool-call hacking exhibits two direct and recurring
manifestations in retrieval-augmented agents trained with format- and
answer-level rewards.

First, the agent’s tool usage becomes poorly coordinated and collapses
to a narrow or repetitive calling pattern.
Instead of adaptively planning tool invocations based on the query
structure and intermediate findings, the policy tends to overuse a
single tool or a fixed call template that reliably satisfies format
constraints.

Second, the agent’s stepwise reasoning or rollout-level answers become
causally disconnected from the retrieved content.
That is, semantically meaningful perturbations to the tool-returned
results do not induce corresponding changes in intermediate reasoning
steps or the final answer.

\begin{figure*}[t]
  \centering
  \includegraphics[width=1\linewidth]{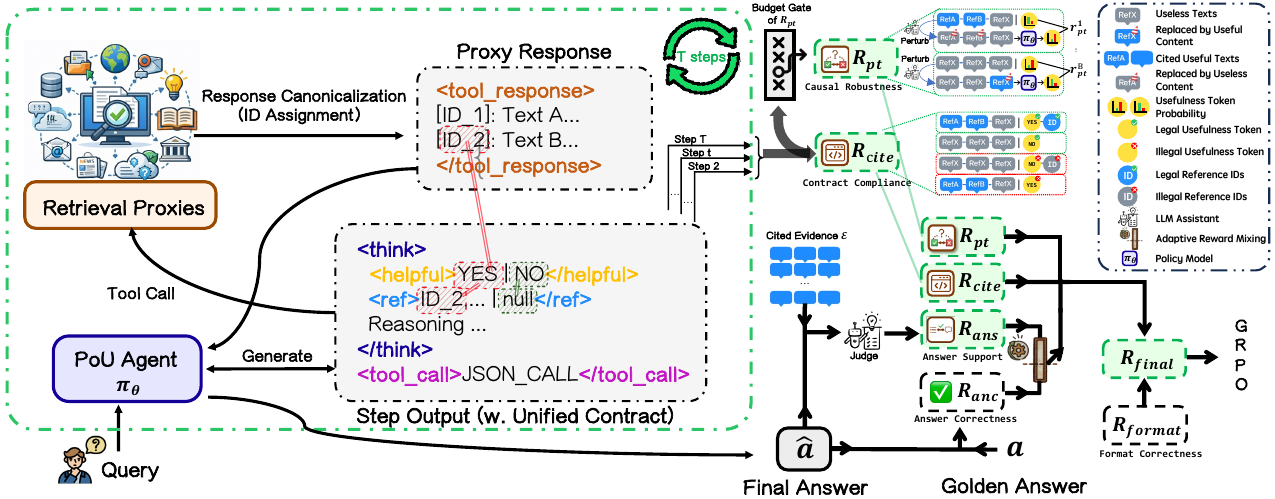}
  \caption{The Proof-of-Use Framework}
  \label{fig:example}
  \vspace{-1em}
\end{figure*}

\label{sec:method}

\subsection{Settings of PoU}\label{sec:setup}

\subsubsection{\textbf{Initialization}}
Each episode begins with a \textbf{user query} $x$ and a predefined \textbf{initial prompt} that provides explicit, structured schemas for each callable tool, including their invocation formats and argument descriptions.  

To simulate realistic multi-source information environments for retrieval–reasoning agents, 
we design and implement four categories of black-box retrieval proxies that collectively capture the diversity of external knowledge access in agentic RAG systems. 
Each proxy corresponds to a distinct knowledge source and interaction modality, enabling the model to reason and retrieve in a unified yet source-specific manner:

\begin{itemize}
    \item \textbf{Web Search Proxy.}  
    This module enables the agent to access open-domain information through keyword-based retrieval.  
    Given a textual query, it returns a structured list of webpage entries, each containing a \texttt{title}, \texttt{URL}, and \texttt{snippet}.  
    For controlled evaluation, the current configuration retrieves a fixed top-$k$ set of documents per query, ensuring reproducible coverage across runs.  
    
    \item \textbf{Web Browsing Proxy.}  
    This proxy performs in-depth inspection of given \texttt{URLs} of retrieved webpages and extracts content relevant to the current query. It returns concise, structured summaries to the reasoning model as evidence for downstream use.

    \item \textbf{Local Search Proxy.}  
    This proxy accesses pre-built, fixed knowledge repositories, typically used in private or professional domains such as academic papers, technical reports, enterprise data, or financial documents.  
    Compared with web search, these repositories feature higher knowledge density and lower noise, making them particularly suitable for deeper domain-specific tasks.
    
    \item \textbf{Knowledge Graph Proxy.}  
    This proxy serves as a interface to structured, graph-based knowledge stores.  
    Rather than employing a query-generation agent that issues executable graph queries, we adopt a semantic KG agent that abstracts subgraphs into concise, natural-language summaries.  
    This choice prioritizes robustness and generality across heterogeneous graph schemas, allowing the proxy to be more compatible with the model’s reasoning process.  
    The KG proxy thus functions as a lightweight retrieval utility supplying structured yet readable context when graph knowledge is required.

\end{itemize}

These tools are uniformly defined within the initialization prompt, allowing the model to understand their expected arguments and formats before the reasoning begins. 

\subsubsection{\textbf{Proxy Response}}
After reasoning, the policy may issue a structured tool call. 
\begin{center}
\texttt{<tool\_call>JSON\_CALL</tool\_call>}
\end{center}
The environment executes the call and returns a \emph{normalized proxy response} with a unified schema:
\begin{center}
\texttt{<tool\_response>REFERENCE\_LIST\_WITH\_ID</tool\_response>}
\end{center}

Here, the globally unique id of each reference is automatically assigned. All four proxies adopt this schema, so the model can cite evidence via \texttt{<ref>ID\_1,ID\_2</ref>} regardless of the source.

\textbf{Granularity.}
For the \emph{Web Search Proxy}, \texttt{ref} items are organized at the \emph{webpage level}, typically including the page title, URL, and a short snippet.
In contrast, the \emph{Web Browsing}, \emph{Local Search}, and \emph{Knowledge Graph} proxies return \texttt{ref} items at the \emph{paragraph (semantic-chunk) level}, enabling finer-grained evidence grounding.

\subsubsection{\textbf{Reasoning Process}}
At the beginning of each iteration, the agent firstly performs reasoning enclosed within a \texttt{<think>} tag.
For the first step, the model generates initial planning and reasoning solely based on the user question.
From the second step onward, each \texttt{<think>} block must begin with a explicit \emph{helpfulness verdict and citation declaration}, which is treated as a \textbf{unified protocol}:
\begin{center}
\texttt{<helpful>yes|no</helpful><ref>id1,id2,...|null</ref>}
\label{eq:protocol}
\end{center}

where \texttt{<ref>} cites the identifiers of retrieved documents from the previous tool call’s structured output.

This mechanism forces grounding each reasoning step on prior observations.
From a probabilistic probing perspective, the hidden state of \texttt{<helpful>} token quantifies the model’s internal judgment of evidence utility, serving as a distributional anchor that aligns the model’s internal activation patterns with the conditional likelihood of evidence-relevant reasoning trajectories. By optimizing the probability mass of reasoning paths causally dependent on retrieved evidence, it suppresses spurious thoughts that diverge from verifiable context.
Meanwhile, the \texttt{<ref>} citations anchors the evidence utility judgment to explicit external sources, establishing a probing-visible linkage between internal representations and external tool outputs. This allows each intermediate conclusion to be traceably grounded in concrete evidence.

Together, the \texttt{<helpful>} and \texttt{<ref>} mechanisms constitute a dual-channel grounding scheme that leverages their interaction as a measurable and learnable interface between probabilistic alignment and structured evidence-grounded reasoning. 

When the model determines it has gathered sufficient evidence, it terminates the loop and emits:
\begin{center}
\texttt{<answer> ... </answer>}
\end{center}
as the final response to the user.

\subsection{Proof-of-Use Rewards}
\subsubsection{Unified Step protocol for Citation Reward}\label{sec:format_reward}

According to the \emph{unified protocol} (denoted in \ref{eq:protocol}) of each step, all proxy responses return dictionaries containing an unique \texttt{ID}, enabling deterministic auditing. We define the \textbf{Cite Reward} $R_{cite}^t$ in step $t$:
\begin{equation}
\begin{split}
R_{cite}^t =
\begin{cases}
+1, & \text{if } \texttt{parse\_ok}\land\texttt{consistency\_ok}\land\texttt{ids\_valid},\\
-1, & \text{otherwise.}
\end{cases}
\end{split}
\label{eq:format_reward}
\end{equation}
\noindent
\textit{Here,}
$\texttt{parse\_ok}=1$ indicates syntactically valid and closed tags;
$\texttt{consistency\_ok}=1$ requires
$(\texttt{helpful=no}\Rightarrow\texttt{ref=null})$ and
$(\texttt{helpful=yes}\Rightarrow\texttt{ref}\neq\texttt{null})$;
$\texttt{ids\_valid}=1$ means all cited IDs exist in the proxy’s return list.

The rollout-level score is computed as the mean over all valid steps: 
\[
R_{cite} = 
\begin{cases}
\frac{\sum_{t=2}^{T} R_{cite}^{t}}{T-1}, & T>1,\\[4pt]
0, & T=1,
\end{cases}
\]
where $T$ is the total number of reasoning steps. 
This ensures that if the model answers directly without invoking any tool ($T{=}1$), no citation reward is given.

This design separates syntax compliance from citation consistency, ensuring that every reasoning step remains both 
\emph{verifiable} and \emph{traceable} to reasoning-evidence alignment.

\subsubsection{Evidence-Sensitive Reward}
\label{sec:perturb}

To evaluate whether the model's decision truly depends on the cited evidence rather than spurious features,
we design a perturbation-based reward that measures the \emph{sensitivity} of the model’s helpfulness prediction $p_\theta(\texttt{yes})$ to evidence change at step $t$:
\begin{equation}
\begin{split}
R^t_{pt} &= s_t \cdot (q' - q),\\
s_t &=
\begin{cases}
-1,& \text{YES case (degrade supportive evidence)},\\
+1,& \text{NO case (inject semantic lure)}.
\end{cases}
\end{split}
\label{eq:pertYes}
\end{equation}

Specifically, for the YES case, we degrade the cited evidence set $S_t$ by replacing their contents with topic-unrelated snippets,
and observe the probability change $\Delta q = q' - q$, where
$q=p_\theta(\texttt{yes}\mid rsp_{\mathrm{real}})$, $q'=p_\theta(\texttt{yes}\mid rsp_{\mathrm{pert}})$, $rsp_{real}$ is the real tool responses and $res_{pert}$ is the real response with partial replacement.
A model that truly relies on the cited evidence should decrease its helpfulness confidence ($\Delta q < 0$)
after the evidence is corrupted. 

For the complementary NO case, we apply the opposite perturbation:
when the model initially predicts the helpfulness token as \texttt{no},
we randomly select one document from the proxy-returned list and
\emph{replace its content with an LLM-generated snippet that is semantically relevant to the query} 
but not necessarily factual.
This synthetic ``semantic lure'' mimics a plausible piece of supportive evidence. A model that truly relies on the cited evidence should increase its helpfulness confidence ($\Delta q > 0$) after lure evidence are inserted. 

Since each rollout may contain multiple steps, the Evidence-Sensitive reward is computed under a fixed budget gate $B$:
\[
B = \min(T{-}1,\, B_{\max},\, \textstyle\sum_t \mathbb{I}[R_{cite}^{(t)}{=}1]),
\]
where we randomly sample $B$ steps from those with $R_{cite}=1$ (from step 2 to $T$) to apply perturbation. We set \(B_{\max}=1\) for main experiments (thus \(B=1\)), and study \(B=2\) in ablations to assess stability–cost trade-offs.

The final rollout-level Evidence-Sensitive reward is defined as the mean over the selected perturbation outcomes:
\begin{equation}
R_{pt}
= \frac{1}{B}
\sum_{t \in \mathcal{S}_B} R_{pt}^{(t)} .
\end{equation}

In practice, by caching prefix KV states and original \texttt{<helpful>} logits, evidence
perturbation is reduced to a tail-only forward pass; see
Appendix~\ref{app:useful_cache}.

\subsubsection{Answer-Support Alignment Reward}
\label{sec:align}
We consider two complementary training signals: 
an \emph{Answer-Support Alignment} reward $R_{ans}$ and a common
\emph{Answer-Correctness} reward $R_{anc}$.

The large action space and delayed rewards make the agent struggle to assign credit properly, often resorting to shortcut guesses rather than grounded reasoning. To overcome the sparse credit assignment and shortcut reasoning issues, PoU directly provides dense and smooth supervision by encouraging answers grounded in retrieved evidence through the Answer-Support alignment objective.

We compute the Answer–Support reward as:
\begin{equation}
\begin{split}
R_{ans} =
\begin{cases}
0, & \text{if } \sum_t \mathbb{I}[R_{cite}^{(t)}{=}1]=0\\
   &\text{or } T{=}1,\\[3pt]
\mathcal{J}(q,\hat{a},\mathcal{E}) \in[0,1], & \text{otherwise.}
\end{cases}
\end{split}
\label{eq:R_ac}
\end{equation}
where $\mathcal{J}$ denotes the external judge model measures the Answer-Support Alignment score of the generated answer $\hat{a}$ based on the cited evidence $\mathcal{E}$. 

Here, we opt to the Faithfulness\footnote{\url{https://docs.ragas.io/en/stable/concepts/metrics/available_metrics/faithfulness/}} metric of Ragas\cite{ragas2024}. And $R_{anc}$ is the LLM-as-Judge\cite{gu2025surveyllmasajudge, seed2025seed15thinkingadvancingsuperbreasoning} score between the predicted and gold answers.

While $R_{anc}$ reflects the final objective, it is typically sparse or unstable in early training stages, making optimization difficult when used alone. To address this issue, we adopt an \textbf{adaptive reward mixing} strategy that enables a smooth curriculum-style transition from $R_{ans}$ to $R_{anc}$ during training:

At each update step $s$, the answer reward for a rollout $i$ is defined as a convex combination:
\begin{equation}
R^{t,i}_a \;=\; \alpha_s\,R_{ans}^{s,i} + (1-\alpha_s)\,R_{anc}^{s,i},
\end{equation}
where $a_s \in (0,1)$ is a shared mixing weight for the entire batch.

To adaptively adjust $\alpha_t$, we track a batch-level estimate of model correctness.
Specifically, let $R_{anc}^{t,i}\in[0,1]$ denote the raw correctness metric computed per rollout.
We first compute the batch mean
\begin{equation}
\hat c_s \;=\; \frac{1}{bs}\sum_{i=1}^{bs} R_{anc}^{s,i},
\end{equation}
and maintain an exponential moving average (EMA)
\begin{equation}
\bar c_s \;=\; \beta \bar c_{s-1} + (1-\beta)\hat c_s,
\qquad \beta\in[0,1).
\end{equation}

The mixing weight is then defined via a monotone logistic gate:
\begin{equation}
\alpha_s \;=\; \sigma\!\big(\kappa(\tau - \bar c_s)\big),
\end{equation}
where $\tau$ denotes a correctness threshold and $\kappa$ controls the sharpness of transition.
As training progresses and $\bar c_s$ increases, $\alpha_s$ decreases smoothly, shifting the optimization focus from evidence consistency to answer correctness.



\subsubsection{Overall Objective}
\label{sec:overall}

The final reward integrates the three signals into a unified rollout-level objective:
\begin{equation}
R_{\mathrm{final}} =
\begin{cases}
\frac{R_{cite} + R_{pt} + R_{a}}{3}, & \text{if format is valid,}\\
-1, & \text{otherwise.}
\end{cases}
\label{eq:R_final}
\end{equation}
\noindent
where the overall format is considered valid only when all basic format such as \texttt{<think>, <tool\_call> and <answer>} pass the syntactic checks.

All components are jointly optimized using \textbf{Group Relative Policy Optimization (GRPO)\cite{shao2024deepseekmathpushinglimitsmathematical}}.

\subsection{Startup Fine-Tuning}
\label{sec:startup}

To initialize a reliable agent under our \textbf{PoU} reasoning framework, 
we generate expert trajectories by a strong LLM (GPT-5). 
These multi-step reasoning trajectories are reformatted into our standardized \emph{interaction schema}, where each turn explicitly follows the PoU protocol.

For every simulated retrieval call, we emulate the output of distinct proxy tools (\texttt{web\_search}, \texttt{local\_search}, \texttt{browser}, \texttt{KG\_search}), each returning paragraph-level evidence with stable identifiers, ensuring deterministic grounding across different information sources.

To guarantee data reliability, we apply a \textbf{two-stage rejection filter} to all generated trajectories. A trajectory is retained only if:
\begin{enumerate}[label=(\roman*)]
    \item Every intermediate step passes the structural and logical validators—
    syntactic correctness (\texttt{parse\_ok}), internal consistency (\texttt{consistency\_ok}), 
    and reference validity (\texttt{ids\_valid});
    \item The reasoning chain contains a non-trivial number of steps ($3 \leq T \leq 10$),
    encouraging multi-hop diversity while excluding degenerate or excessively long trajectories.
\end{enumerate}

The resulting dataset thus consists of \emph{verifiable, well-grounded reasoning episodes}
that adhere to our explicit stepwise protocol.
We fine-tune the base model on this curated corpus to \textbf{cold-start the PoU agent}.

\section{Experiments}
\subsection{Datasets and Metrics}
\label{sec:datasets}

We evaluate \textbf{PoU} on a mixed \emph{in-/out-of-domain (ID/OOD)} benchmark suite designed to probe open-domain QA under distributional shifts.

\paragraph{In-Distribution (ID)}
Since our training data are drawn exclusively from \textbf{HotpotQA}~\cite{yang2018hotpotqadatasetdiverseexplainable} and \textbf{2WikiMultihopQA}~\cite{xanh2020_2wikimultihop},
we report ID performance on their official development splits.

\paragraph{Out-of-Distribution (OOD)}
To assess generalization beyond training domains, we evaluate on five additional datasets with distinct question styles and evidence structures:
\textbf{Natural Questions (NQ)}~\cite{47761},
\textbf{TriviaQA (TQ)}~\cite{joshi2017triviaqalargescaledistantly},
\textbf{MuSiQue}~\cite{trivedi2022musiquemultihopquestionssinglehop},
\textbf{PopQA}~\cite{mallen2023trustlanguagemodelsinvestigating},
and \textbf{Bamboogle}~\cite{press2023measuringnarrowingcompositionalitygap}.

For comparability across datasets, we uniformly sample $512$ instances from each dev set of NQ, TQ, HotpotQA, 2Wiki, MuSiQue, and PopQA (using a fixed global seed),
and include all $125$ items from the Bamboogle dev set.
This yields a total of $3{,}197$ evaluation questions, maintaining balanced variance across datasets
while avoiding confounds from dataset size.
Only HotpotQA and 2Wiki are used for training; all others serve as strict OOD evaluation.

\subsection{Baselines}
\label{sec:baselines}

We compare PoU with representative \textbf{deep-research agents}
covering retrieval–reasoning RAG, RL-based search, and multi-tool agent systems.

\paragraph{\textbf{(1) Retrieval–Reasoning RAG.}}
\textbf{Routing RAG} dynamically selects retrievers (e.g., corpus, web, KG)
based on query type and reasoning context,
while \textbf{All-in-One RAG} fuses multiple retrievers into a unified interface
for joint multi-source reasoning.
Both follow the \textbf{ReAct}~\cite{yao2023reactsynergizingreasoningacting} framework
with three reasoning–retrieval turns.

\paragraph{\textbf{(2) RL-based Search Agents.}}
\textbf{Search-o1}~\cite{Li2025Searcho1AS},
\textbf{Search-R1}~\cite{jin2025searchr1trainingllmsreason},
and \textbf{R1-Searcher}~\cite{song2025r1searcherincentivizingsearchcapability}
perform multi-hop search via RL-guided sub-query generation,
primarily over Wikipedia, followed by evidence summarization.

\paragraph{\textbf{(3) Multi-Tool Agents.}}
\textbf{DeepResearcher}~\cite{zheng2025deepresearcherscalingdeepresearch} and
\textbf{Chain-of-Agent (CoA)}~\cite{li2025chainofagentsendtoendagentfoundation}
represent strong recent multi-tool agents operating over heterogeneous tool environments for deep research. \textbf{CoA} is trained by distilling trajectories from strong multi-agent systems and further refined with agentic reinforcement learning. \textbf{DeepResearcher} is an end-to-end reinforcement learning framework that trains
LLM-based research agents in real-world web environments by learning to select and
coordinate multiple web tools for search and browsing, using standard format and
answer-correctness rewards, as in RL-based search agent baselines.
For fair comparison, we train DeepResearcher from scratch using the same external tool set as PoU (denoted as \textbf{DeepResearcher+}), enabling evaluation under identical access to external knowledge sources.

\paragraph{\textbf{(4) Distillation-Only PoU (SFT).}}
To further justify the necessity of explicit process-level RL supervision of PoU, we include an SFT-only baseline (\textbf{PoU-SFT}) trained on PoU-formatted expert trajectories, using the same model, tools, and data as PoU but without any RL rewards, thereby disentangling the effects of explicit process-level RL supervision
from framework design and teacher–student distillation.

\subsection{Implementation Details}
We use GPT-4o-mini as the external judge model for evaluating answer–evidence alignment and factual consistency.
All other agentic modules (browser, KG proxy, perturbations LLM) are implemented with Qwen-3-8B, deployed via vLLM inference servers. The training backbone model is Qwen-2.5-7B-Instruct.
Prior to reinforcement learning, supervised fine-tuning is performed with a batch size = 256 for 3 epochs.
For reinforcement learning, we adopt VERL with the GRPO algorithm. Each update uses 8 parallel rollouts, distributed across 32 × A100 GPUs. The $R_{pt}$ budget $B$ is 1.
The maximum reasoning step is set to 10, and each retrieval call returns the top-5 documents or chunks per query.
The KG proxy retrieves one-hop subgraph information from Wikidata and WikiPedia by ToG-2\cite{ma2025thinkongraph} system, providing semantically condensed neighborhood summaries as context. In the OOT experiments, we additionally introduce a local biochemical Proxy, which is built from a biochemical corpus derived from PubMed and serves as a local retrieval interface, serving as the fixed local knowledge base for domain-specific retrieval. The local\_proxy is turn-off for non-medical domain tasks.

\subsection{Main Results}
\begin{table}[t]
\centering
\caption{In-domain evaluation on HotpotQA and 2Wiki datasets. All results are reported with F1 and LLM-based evaluation (LM) metrics. 
Bold numbers indicate the best performance in each column. Underlined numbers indicate the second best.}
\label{tab:benchmark_in}
\renewcommand{\arraystretch}{1.2}
\setlength{\tabcolsep}{7pt}
\small
\begin{tabular}{lc|cc|cc}
\toprule
\textbf{Model} & \textbf{Env.} &
\multicolumn{2}{c|}{\textbf{HotpotQA}} &
\multicolumn{2}{c}{\textbf{2Wiki}} \\
\cmidrule(lr){3-4}\cmidrule(lr){5-6}
 & & F1 & LM & F1 & LM \\
\midrule
\multicolumn{6}{l}{\textit{\textbf{RAG}}} \\
\textbf{Routing RAG} & Multi-source & 30.8 & 36.0 & 14.4 & 18.2 \\
\textbf{All-in-one RAG} & Multi-source & 34.5 & 40.2 & 23.2 & 26.4 \\
\textbf{Search-o1}* & Local & 31.6 & 40.8 & 28.6 & 32.8 \\
\textbf{Search-o1} & Web Search & 33.0 & 42.4 & 30.9 & 37.7 \\
\midrule
\multicolumn{6}{l}{\textit{RL Deep Research}} \\
\textbf{Search-R1-base} & Local & \textbf{55.9} & 63.0 & 44.6 & 47.9 \\
\textbf{Search-R1-instruct} & Local & 45.7 & 52.5 & 43.4 & 48.8 \\
\textbf{R1-Searcher} & Web Search & 44.8 & 53.1 & 59.4 & 65.8 \\
\textbf{CoA} & Multi-source & — & 43.9 & — & 50.2 \\
\textbf{DeepResearcher} & Web Search & 52.8 & \underline{64.3} & 59.7 & 66.6 \\
\textbf{DeepResearcher+} & Multi-source & 50.3 & 61.7 & \underline{61.2} & \underline{67.4} \\
\textbf{PoU-SFT }& Multi-source & 42.2 & 45.5 & 37.6 & 41.9 \\
\midrule
\multicolumn{6}{l}{\textit{\textbf{Ours}}} \\
\textbf{PoU} & Multi-source & \underline{54.9} & \textbf{67.2} & \textbf{68.3} & \textbf{78.0} \\
\bottomrule
\end{tabular}

\end{table}

\begin{table*}[t]
\centering
\caption{Performance comparison on five out-of-domain QA benchmarks: \textbf{NQ}, \textbf{TQ}, \textbf{MuSiQue}, \textbf{Bamboogle}, and \textbf{PopQA}. 
All results are reported with F1 and LLM-based evaluation (LM) metrics. 
Bold numbers indicate the best performance in each column. Underlined numbers indicate the second best.}
\label{tab:ood_benchmarks}
\renewcommand{\arraystretch}{1.2}
\setlength{\tabcolsep}{5.5pt}
\small
\begin{tabular}{lc|cc|cc|cc|cc|cc}
\toprule
\textbf{Method} & \textbf{Env.} &
\multicolumn{2}{c|}{\textbf{NQ}} &
\multicolumn{2}{c|}{\textbf{TQ}} &
\multicolumn{2}{c|}{\textbf{MuSiQue}} &
\multicolumn{2}{c|}{\textbf{Bamboogle}} &
\multicolumn{2}{c}{\textbf{PopQA}} \\
\cmidrule(lr){3-4} \cmidrule(lr){5-6} \cmidrule(lr){7-8} \cmidrule(lr){9-10} \cmidrule(lr){11-12}
 & & F1 & LM & F1 & LM & F1 & LM & F1 & LM & F1 & LM \\
\midrule
\multicolumn{12}{l}{\textit{\textbf{RAG}}} \\
\textbf{Routing RAG} & Multi-source & 34.8 & 42.9 & 51.0 & 59.6 & 6.5 & 8.2 & 17.0 & 17.4 & 35.1 & 39.9 \\
\textbf{All-in-one RAG} & Multi-source & 37.5 & 48.3 & 60.2 & 68.8 & 9.7 & 11.8 & 23.5 & 25.1 & 46.9 & 48.8 \\
\textbf{Search-o1}* & Local & 34.5 & 57.4 & 52.6 & 61.1 & 16.8 & 21.3 & 35.8 & 38.4 & 36.9 & 42.4 \\
\textbf{Search-o1} & Web Search & 32.4 & 55.1 & 58.9 & 69.5 & 14.7 & 19.7 & 46.6 & 53.6 & 38.3 & 43.4 \\
\midrule
\multicolumn{12}{l}{\textit{\textbf{RL Deep Research}}} \\
\textbf{Search-R1-base} & Local & \underline{45.4} & 60.0 & 71.9 & 76.2 & 26.7 & 27.5 & 56.5 & 57.6 & 43.2 & 47.0 \\
\textbf{Search-R1-instruct} & Local & 33.1 & 49.6 & 44.7 & 49.2 & 26.5 & 28.3 & 45.0 & 47.2 & 43.0 & 44.5 \\
\textbf{R1-Searcher} & Web Search & 35.4 & 52.3 & 73.1 & 79.1 & 22.8 & 25.6 & 64.8 & 65.6 & 42.7 & 43.4 \\
\textbf{CoA} & Multi-source & — & 43.9 & — & 63.3 & — & 22.3 & — & 49.6 & — & 46.5 \\
\textbf{DeepResearcher} & Web Search & 39.6 & \underline{61.9} & \textbf{78.4} & \underline{85.0} & \underline{27.1} & \underline{29.3} & \textbf{71.0} & \underline{72.8} & 48.5 & 52.7 \\
\textbf{DeepResearcher}\textbf{+} & Multi-source & 37.9 & 58.7 & 76.0 & 82.1 & 23.0& 26.5 & 68.9& 70.4  & \underline{51.1} & \underline{54.5} \\
\textbf{PoU-SFT} & Multi-source & 35.9 & 56.0 & 51.3 & 60.6 & 11.2& 12.4 & 33.3& 35.2 & 37.2 & 41.4 \\
\midrule
\multicolumn{12}{l}{\textit{\textbf{Ours}}} \\
\textbf{PoU} & Multi-source & \textbf{45.5} & \textbf{67.2} & \underline{77.6} & \textbf{89.6} & \textbf{29.0} & \textbf{34.1} & \underline{69.5} & \textbf{74.6} & \textbf{55.7} & \textbf{61.2} \\
\bottomrule
\end{tabular}
\end{table*}
Across all seven datasets—covering both in-domain (HotpotQA, 2Wiki) and out-of-domain (NQ, TQ, MuSiQue, Bamboogle, PopQA) benchmarks—our proposed \textbf{Proof-of-Usefulness (PoU)} agent consistently outperforms prior RAG and RL-based baselines.
Notably, these gains extend beyond the training distribution: on all out-of-domain benchmarks, PoU achieves clear improvements in both factual F1 and LLM-as-judge (LM) metrics, demonstrating strong generalization ability.

Importantly, PoU also surpasses baselines on datasets such as NQ and TQ, which constitute in-domain data for the \textbf{DeepResearcher} baseline.
This indicates that PoU’s improvements do not stem from memorizing domain-specific knowledge, but from stronger \emph{evidence--reasoning alignment}.
The model learns to effectively utilize retrieved evidence and to transfer its reasoning strategies to novel domains, capturing transferable reasoning competence rather than dataset-specific experience.

Results from \textbf{DeepResearcher+} further show that naively expanding the action space by introducing additional information sources can degrade standard RL agents: they fail to exploit the added knowledge and suffer performance drops.
This highlights the necessity of specialized training strategies for multi-source tool use.
Moreover, \textbf{PoU-SFT}, despite being trained on LLM-guided expert trajectories, exhibits limited capability and degrades more severely on OOD tasks, suggesting that PoU’s gains do not arise from LLM-based rewards alone but from evidence-grounded training that teaches decision-making logic rather than simple imitation.

We also observe that \textbf{All-in-One RAG} consistently outperforms \textbf{Routing RAG} across all evaluation sets, corroborating findings from \emph{RAG in the Wild} that naive or heuristic routing—without explicit optimization or supervision—is often inefficient or even counterproductive.
While routing appears intuitively appealing by narrowing retrieval scope and reducing search cost, it risks prematurely discarding relevant information.

In our setting, the PoU agent operates in a multi-source environment where routing becomes an additional latent decision variable.
Compared with traditional deep-research agents, this introduces a new layer of complexity: at each reasoning step, the agent must decide not only \emph{what} to retrieve but also \emph{where} to retrieve from.
As the number of information sources grows, this expanded action space leads to inefficient exploration unless the routing policy is explicitly constrained.
PoU addresses this challenge through structured reasoning and protocol-based training: by explicitly enforcing citation validity and reasoning consistency at every step, PoU regularizes the search process and prevents divergence in the enlarged action space.
As a result, PoU can effectively leverage multi-source evidence without the instability and inefficiency characteristic of naive routing-based systems, yielding \textbf{superior performance and generalization in multi-source environments}.

\subsection{Ablation Study}
\subsubsection{Rewards Ablation}
\label{sec:ablation}

\begin{figure}[t]
    \centering
    \includegraphics[width=1\linewidth]{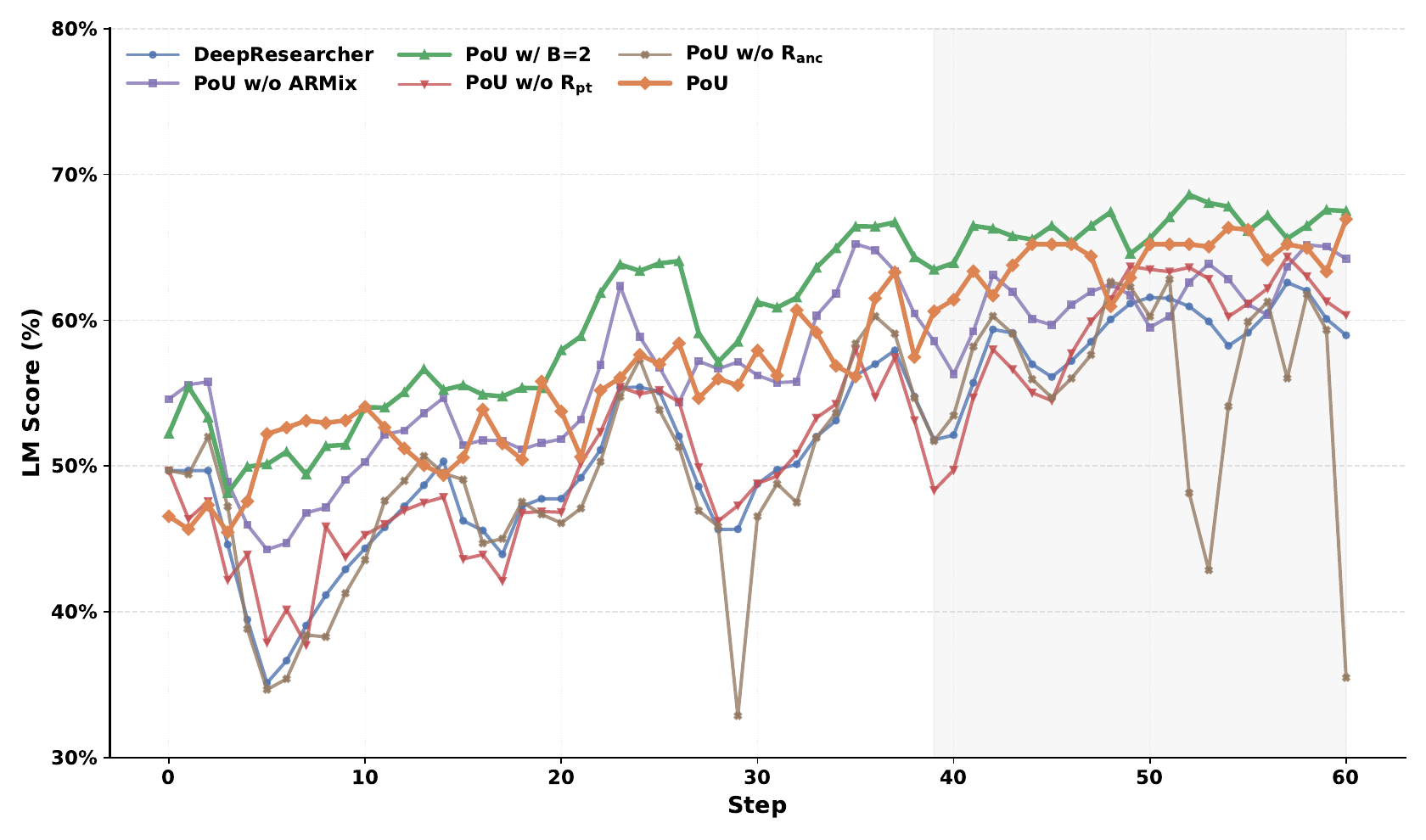}
    \caption{
    \textbf{Ablation study of PoU.}
    Comparison under different reward configurations:
    default perturbation budget $B{=}1$, extended $B{=}2$, and variants removing $R_{pt}$ or $R_{ans}$.}
    \label{fig:pou_ablation}
    \vspace{-1.5em}
\end{figure}

Figure~\ref{fig:pou_ablation} presents the ablation study of PoU under different reward configurations.
We observe that increasing the perturbation budget to $B{=}2$ leads to a more stable training trajectory and a higher final performance.
This confirms the effectiveness of the perturbation-based reward $R_{pt}$:
by actively perturbing the cited evidence, the model learns to rely on genuinely supportive information
rather than superficially acknowledging tool responses. In addition, after removing \emph{adaptive reward mixing} (i.e., using a static average of $R_{ans}$ and $R_{anc}$), compared to PoU, the learning curve shows a slower ascent in the early stage and a lower performance ceiling at convergence. This behavior reflects the stage-dependent role of reward signals. In the early phase, the model relies more heavily on $R_{\text{ans}}$ to accelerate learning by guiding coarse-grained process optimization. As training progresses and evidence--reasoning consistency is largely established, the model increasingly benefits
from stronger answer-level supervision. This observation is well aligned with the original motivation of
\emph{adaptive reward mixing}. Furthermore, removing the answer–citation alignment term $R_{\text{ans}}$ leads to highly unstable training dynamics and frequent collapse in later stages.
This suggests that explicitly linking the final answer with intermediate evidence provides an essential smoothing and grounding signal for policy optimization.

\subsubsection{Out-of-domain-tool (OOT) and negative tool (NT) Experiments}
To evaluate generalization under simultaneous domain and tool-set shifts, we conduct additional experiments in an \textbf{out-of-domain (OOD)} and \textbf{out-of-domain-tool (OOT)} setting.
Specifically, we introduce tools that are never observed during training (a biomedical local search proxy (Local\_Bio)).
In addition, we introduce a \textbf{negative tool (NT)}, which is deliberately task-irrelevant, to reflect realistic scenarios where certain tools are inapplicable and to assess the model’s robustness to interference as well as its ability to select appropriate tools.
Experiments are conducted on two biomedical QA benchmarks, \textbf{BioASQ}~\cite{krithara2023bioasq} and \textbf{PQArefEval}~\cite{bašaragin2024knowthatteachinggenerative}.

\begin{table*}[t]
\centering
\caption{OOT evaluation on biomedical QA datasets (\textbf{BioASQ}, \textbf{PQArefEval}). ``w/ NT'' denotes the inclusion of an additional negative tool.}
\label{tab:biomed_ood}
\renewcommand{\arraystretch}{1.2}
\setlength{\tabcolsep}{6pt}
\small
\begin{tabular}{lccccccc}
\toprule
\textbf{Dataset} & \textbf{Zero-shot} & \textbf{RAG} & \textbf{DeepResearcher+} & \textbf{DeepResearcher} &\textbf{DeepResearcher (w/ NT)} & \textbf{PoU} & \textbf{PoU (w/ NT)}\\
\midrule
\textbf{BioASQ} & 13.5 & 25.8 & 37.1 & 38.9 & 31.8& \textbf{45.6} & 43.9\\
\textbf{PQArefEval} & 18.4 & 31.5 & 42.4 & 42.1 & 37.6&\textbf{48.7} & 48.3\\
\bottomrule
\end{tabular}·
\vspace{-1em}
\end{table*}

As shown in Table~\ref{tab:biomed_ood}, DeepResearcher+, despite being trained with an expanded toolset, performs consistently worse than its original 2-tool counterpart (DeepResearcher) not only in OOD settings but also under the more challenging \textbf{OOT} regime across both biomedical datasets. Moreover, when an additional negative tool (NT) is introduced, the performance of DeepResearcher drops substantially, indicating a strong sensitivity to task-irrelevant tools.
While PoU model achieves the best performance by a clear margin, demonstrating strong domain-level and tool-level generalization, while remaining robust even when the negative tool (NT) is introduced.

This result reveals an important phenomenon: as the number of available tools increases, the agent’s action space grows combinatorially.
Without explicit process-level supervision, such an enlarged search space can easily lead to unstable policies or degenerate reasoning behaviors.

In contrast, through explicit evidence–reasoning-answer alignment training, the PoU training paradigm stabilizes learning under both domain and tool distribution shifts, substantially mitigating tool-call hacking.
Crucially, PoU does not explicitly optimize for tool adaptation or robustness to task-irrelevant tools.
Instead, by regulate a tight dependency between evidence usage, intermediate reasoning, and final answers, the training objective implicitly biases the trajectory toward tools that consistently support verifiable reasoning.
\textbf{As a result, robustness to unseen or task-irrelevant tools emerges as a extra bonus from PoU.}

\subsubsection{Tool-Call Behavior Analysis}
\label{sec:tool_call_ratio}
\begin{figure}[t]
    \centering
    \includegraphics[width=\linewidth]{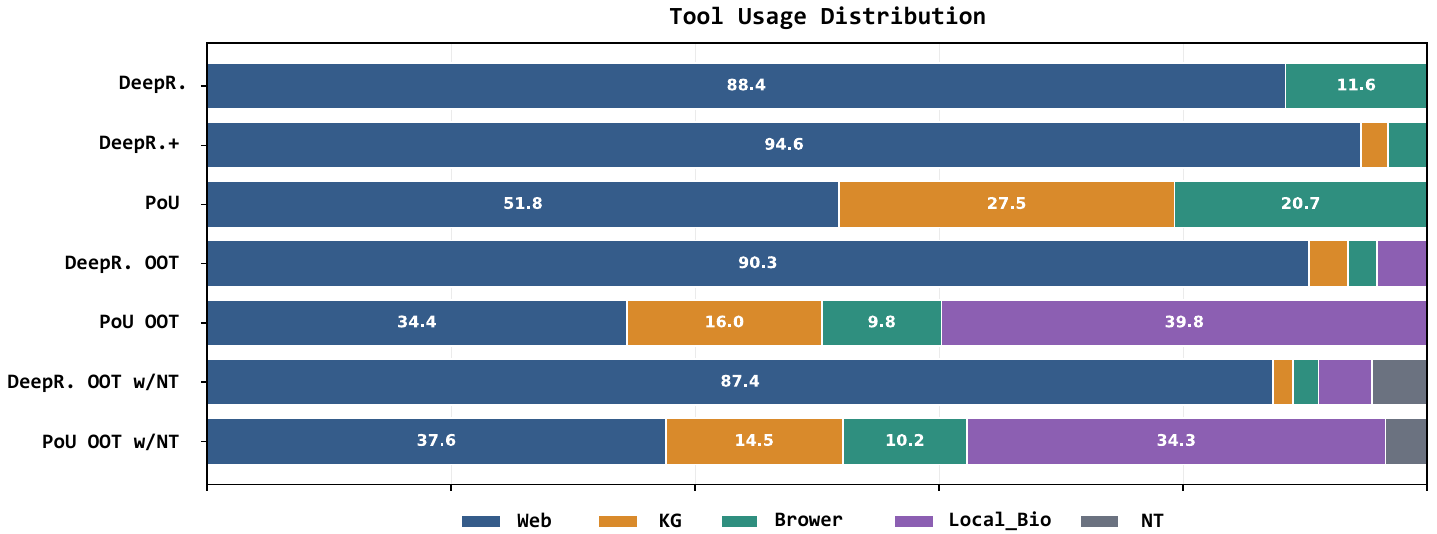}
    \caption{Tool-call ratio (\%) distributions across models and settings. For visual clarity, values below 6\% are not shown. For brevity, \textbf{DeepR.} denotes \textbf{DeepResearcher}.
    }
    \label{fig:usg}
\end{figure}

Building on Figure~\ref{fig:example1}, which illustrates divergent tool-entropy dynamics between PoU and the baseline during training, we further examine how agents adapt their tool-call patterns under different task settings and tool environments. We uniformly sample 1,000 rollouts from each of seven open-domain test sets (HotpotQA, 2Wiki, NQ, TQ, …) and two out-of-tool-distribution (OOT) benchmarks (BioASQ, PQArefEval), and report the resulting tool-call ratio distribution across tools in Figure~\ref{fig:usg}.

With only the format and answer rewards (as in the original DeepResearcher setup), the agent tends to overuse available tools once the tool set expands.
After trained with one more tool proxy, DeepResearcher+ exhibits a pronounced bias toward redundant Web Search calls (94.6\%) and a near disappearance of Web Browser usage (3.2\%).
This behavior indicates a collapse of tool specialization: despite the intended search–browse hierarchy, the agent fails to differentiate tool roles and repeatedly queries the Web Search proxy. This indicates not merely a preference imbalance, but a failure to learn functional differentiation among tools. Similarly, under OOT and NT settings, DeepResearcher maintains a highly polarized and stable tool-call distribution, reflecting a rigid tool usage pattern that persists despite changes in task relevance.
From a training perspective, this behavior arises because outcome-level rewards dominate optimization. Once a particular tool consistently yields non-zero rewards during early training, the policy is reinforced to repeatedly invoke it, while alternative tools that lack step-level corrective signals are gradually suppressed. As a result, the agent converges to a brittle retrieval strategy that fails to adapt under tool-set or distribution shifts. \textbf{Such retrieval actions suggests severe Tool-call Hacking of normal format-answer supervision again.} 

In contrast, the PoU agent exhibits a more balanced tool-call distribution.
\textbf{Importantly, beyond the robustness observed in OOT and NT settings, this balance reflects a second emergent property: a non-static and context-sensitive tool allocation that varies across tasks and tool environments, indicating flexible adaptation of heterogeneous retrieval utilities.}

\subsubsection{Extreme Case}
To expose tool-call hacking and evaluate resistance to spurious tool response, we intervene in tool-call by randomly replacing the content of the corresponding \texttt{<tool\_response>} with the token \texttt{``content''}, while preserving the response structure. Due to space constraints, Table~\ref{tab:step_think_comparison} only presents the
\texttt{<think>} component of a single reasoning step under spurious tool responses.
Nevertheless, this simplified illustration faithfully reflects the behavior patterns
we consistently observe in full rollouts.
In practice, when the tool response content is corrupted while preserving its structure,
\textbf{DeepResearcher} often produces seemingly coherent intermediate reasoning that
implicitly assumes the existence of valid evidence, despite the absence of any usable
information in the tool output.
In contrast, \textbf{PoU} explicitly recognizes the lack of supporting evidence at the
step level, marks the tool call as unhelpful, and proactively resorts to alternative tools.
This contrast is robust across queries and steps, and the extreme case shown here serves
as a minimal yet representative example of the underlying phenomenon.
\begin{table}[t]
\centering
\small
\setlength{\tabcolsep}{5pt}
\renewcommand{\arraystretch}{1.15}
\caption{An Extreme Case}
\vspace{-0.5em}
\label{tab:step_think_comparison}

\noindent\text{Observation:}\texttt{<tool\_response> content </tool\_response>}

\vspace{0.2em}
\begin{tabularx}{\linewidth}{p{2.7cm} X}
\toprule
\textbf{Method} &  \\
\midrule
\textbf{DeepResearcher} &
\texttt{<think>} \\
& I found the information about ``xxx''. \\
& Now I will check ``xxx''. \\
& \texttt{</think>} \\
\midrule
\textbf{PoU} &
\texttt{<think>} \\
& \texttt{<helpful>}no\texttt{</helpful>} \\
& \texttt{<ref>}null\texttt{</ref>} \\
& No relevant information found. \\
& I will try another tool. \\
& \texttt{</think>} \\
\bottomrule
\end{tabularx}
\vspace{-2em}
\end{table}

\section{Conclusion}

In this work, we identify \emph{tool-call hacking} as a subtle yet critical failure mode in RL-trained deep research agents, arising from the weak observability of evidence usage under outcome-dominated supervision. To address this issue, we propose Proof-of-Use (PoU), an evidence-grounded training framework that explicitly enforces causal dependencies between retrieval, reasoning, and final answers. Extensive experiments across diverse tasks and tool environments demonstrate that PoU effectively mitigates tool-call hacking and yields strong performance. Beyond this, PoU exhibits notable emergent properties, including robustness and adaptive tool-usage behaviors under domain and tool shifts, despite not being explicitly optimized for tool adaptation.


\bibliographystyle{ACM-Reference-Format}
\bibliography{sample-base}

\appendix

\end{document}